%% file: main.tex
\documentclass[runningheads]{llncs}

\usepackage[sort]{cite}
\usepackage[T1]{fontenc}

\usepackage{graphicx,verbatim}
\usepackage{subcaption}

\usepackage{booktabs}
\usepackage{makecell}
\usepackage{multicol}
\usepackage{multirow}

\input{notation}

\begin{document}

\title{Test-Time-Scaling for Zero-Shot Diagnosis with Visual-Language Reasoning}

\author{Ji Young Byun\inst{1} \and Young-Jin Park\inst{2} \and Navid Azizan\inst{2} \and Rama Chellappa\inst{1}} 

\institute{Johns Hopkins University, Baltimore MD, USA \and Massachusetts Institute of Technology, Cambridge MA, USA \\
    \email{\{jbyun13, rchella4\}@jhu.edu, \{youngp, azizan\}@mit.edu}}

\maketitle

\begin{abstract}
As a cornerstone of patient care, clinical decision-making significantly influences patient outcomes and can be enhanced by large language models (LLMs). Although LLMs have demonstrated remarkable performance, their application to visual question answering in medical imaging, particularly for reasoning-based diagnosis, remains largely unexplored. Furthermore, supervised fine-tuning for reasoning tasks is largely impractical due to limited data availability and high annotation costs.
In this work, we introduce a zero-shot framework for reliable medical image diagnosis that enhances the reasoning capabilities of LLMs in clinical settings through test-time scaling. 
Given a medical image and a textual prompt, a vision-language model processes a medical image along with a corresponding textual prompt to generate multiple descriptions or interpretations of visual features. These interpretations are then fed to an LLM, where a test-time scaling strategy consolidates multiple candidate outputs into a reliable final diagnosis. We evaluate our approach across various medical imaging modalities---including radiology, ophthalmology, and histopathology---and demonstrate that the proposed test-time scaling strategy enhances diagnostic accuracy for both our and baseline methods. Additionally, we provide an empirical analysis showing that the proposed approach, which allows unbiased prompting in the first stage, improves the reliability of LLM-generated diagnoses and enhances classification accuracy.


\end{abstract}

\section{Introduction}

Addressing the complexity of medical tasks requires a comprehensive reasoning process beyond multiple-choice responses, such as MedQA \cite{jin2021disease} and PubMedQA \cite{jin2019pubmedqa}. 
Although their near-human performance across various applications has enabled their integration into multiple healthcare contexts, text-only models can fail to fully represent the multimodal nature of clinical practice, a critical factor for precise decision-making. In light of this, incorporating Vision Language Models (VLMs)---which take images and textual prompts to generate textual outputs---is a key enabler for achieving a multifaceted approach employed by clinicians. Furthermore, several studies have validated the approach of treating medical image diagnosis as a form of visual question answering (VQA) using fine-tuned VLMs \cite{li2023llava,tu2024medpalm,moor2023medflamingo}.

Conventional data-driven deep learning approaches parameterize a model with learnable parameters and train it on a dataset of image--label pairs. Such models are often treated as \emph{black-box} predictors: they provide a final classification result but lack transparency, making the diagnostic reasoning process non-interpretable. Considering the safety-critical nature of medicine and the risk that generated texts may not align with clinical standards, evaluating these models is necessary to assess progress and minimize potential harms \cite{johnson2023assessing}.

To address this, explicit text-based multi-step reasoning methods, such as chain-of-thought \cite{wei2022chain, temsah2024openai}, can promote interpretability. 
By integrating a step-by-step breakdown of intermediate reasoning steps that culminate in the final answer, multi-step reasoning methods enhance interpretability and problem-solving capabilities. While they have shown effectiveness in domains such as arithmetic and symbolic reasoning, their application to multimodal medical tasks remains limited due to data scarcity and the high cost of human annotations. These constraints underscore the need for methods that leverage LLMs without extensive fine-tuning, making zero-shot approaches particularly appealing.

Zero-shot prediction with (small-size) LLMs, however, often yields suboptimal performance.
To address this, a recently proposed inference paradigm known as \emph{test-time scaling} (TTS)  has shown promise in enhancing their effectiveness.
Rather than relying on a single-pass decoding, the key idea of TTS is to sample multiple candidate outputs and aggregate them \cite{yao2023tree,snell2024scaling}.
These approaches, ranging from majority voting to verifier-based selection \cite{cobbe2021training, uesato2022solving, want2024shepherd, lightman2024lets}, have demonstrated strong performance in tasks requiring complex reasoning, such as mathematics. However, they face challenges in medical applications: majority voting can amplify noise when the LLM is poorly calibrated, common in underexplored tasks like disease diagnosis, while verifier-based methods demand extensive labeled data for training. Motivated by these limitations, we explore simple yet effective TTS strategies that enable reliable zero-shot medical diagnosis without requiring additional supervision.

To this end, we propose a zero-shot disease classification framework aimed at improving LLMs' zero-shot capabilities in clinical applications. Initially, the VLM receives a medical image and generates a set of interpretations of the image. This set of information is then passed on to LLMs, with test-time scaling applied to determine the final diagnosis.
Our evaluation investigates LLM capabilities across diverse medical image modalities, including radiology, ophthalmology, and histopathology images.

Our key contributions are summarized as follows:
\begin{enumerate}
    \item We leverage a multi-stage reasoning approach, facilitating disease diagnosis for multimodal LLMs.
    \item We enhance zero-shot accuracy by presenting a test-time scaling method that effectively integrates multiple visual descriptions extracted from VLMs.
    \item We provide an empirical analysis of how biased prompting undermines the reliability of LLM-generated answers and explore a mitigation strategy.
\end{enumerate}

\begin{figure}[t!]
    \centering
    \begin{subfigure}[b]{0.62\textwidth}
        \centering
        \includegraphics[width=\textwidth]{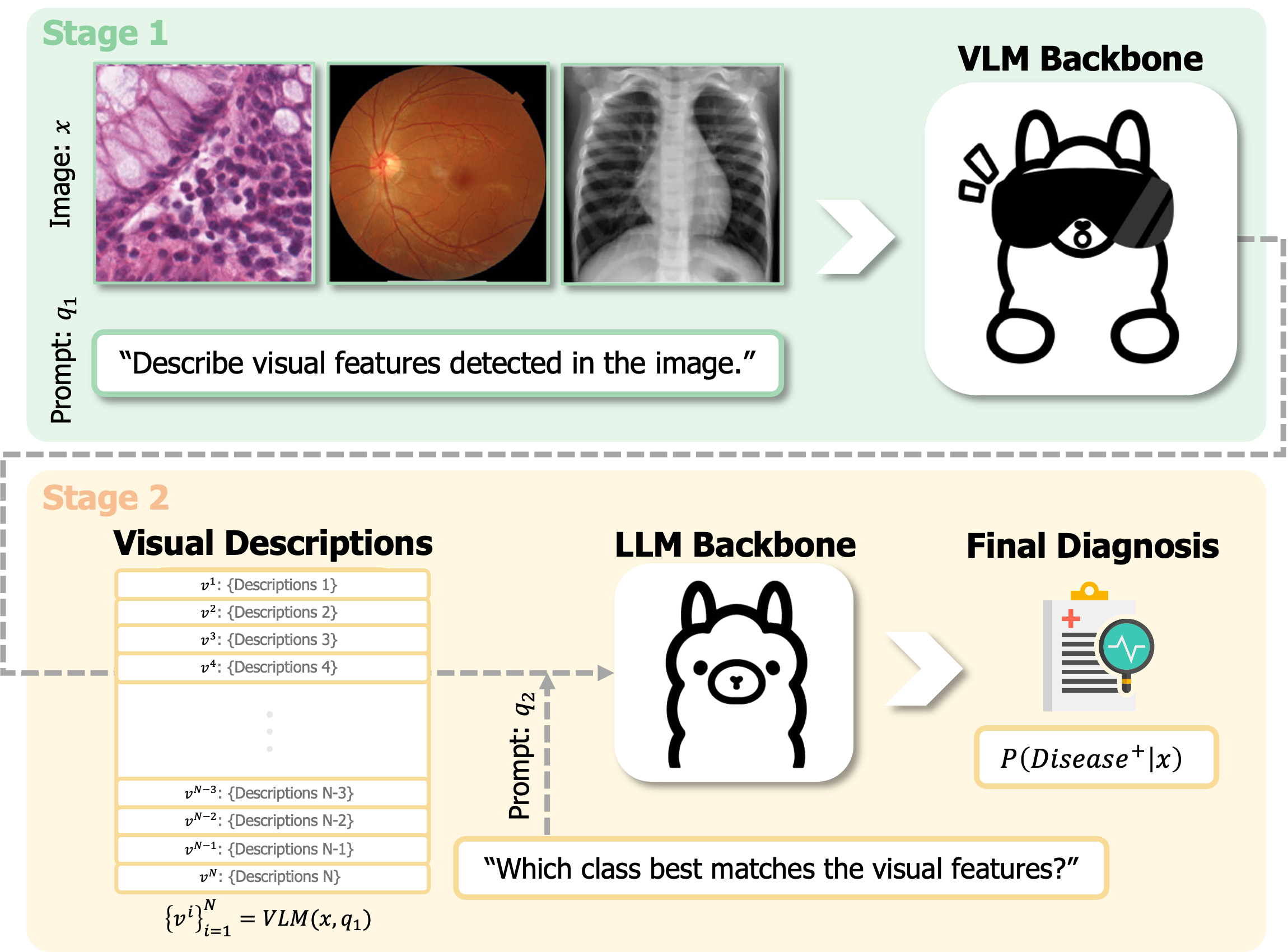}
        \caption{}
    \end{subfigure}
    \begin{subfigure}[b]{0.35\textwidth}
        \centering
        \includegraphics[width=\textwidth]{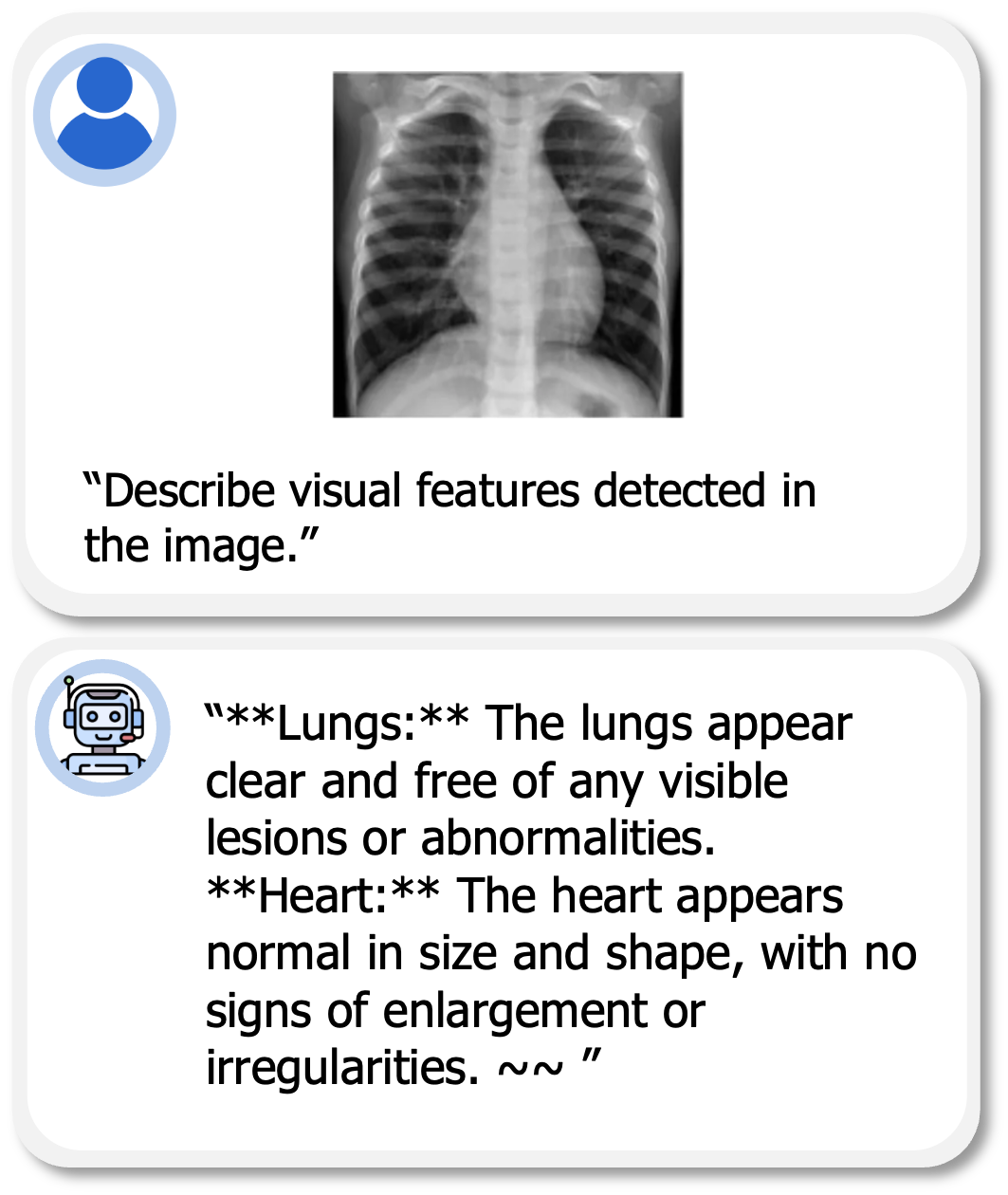} 
        \caption{}
    \end{subfigure}
    \caption{Graphical illustration of our proposed test-time-scaled reasoning framework for reliable zero-shot medical image diagnosis. Panel (a) details Stage 1, in which the VLM receives an image and a text prompt to generate $N$ visual description samples, with Stage 2 applying a test-time scaling technique to determine the final diagnostic probability. Panel (b) shows a representative example of a healthy subject's chest X-ray paired with textual prompt and one of the $N$ generated visual descriptions from VLM in Stage 1.}
    \label{fig:framework}
\end{figure}

\section{Method}

\subsection{Medical Image Diagnosis as Visual Question Answering}

\subsubsection{Problem Statement.}
We consider a medical image $\bx$ of the patient together with its corresponding ground-truth diagnosis label $y$. 
The objective of automated medical image diagnosis is to learn a model $f$ that approximates the posterior probability of the patient having a certain disease: $p(y \mid \bx)$.

\subsubsection{Baselines: Zero-shot Question Answering (QA).}
Consider a VLM (e.g., Llama 3.2-Vision instruction-tuned model) that takes a textual prompt $q$ and an image $\bx$ and produces an answer $a$, formally written as $a = f(\bx, q)$. 
A straightforward way to use such a VLM for medical image diagnosis is to directly request a prediction of the target diagnosis. 
For instance, we can set 
$q \leftarrow$ \txt{Given a pediatric chest X-ray image, classify it as 0 (normal) or 1 (pneumonia).} 
We refer to this approach as \emph{zero-shot} QA.

An alternative is to prompt the model to provide a step-by-step explanation before giving the final answer. 
This can be done by adding a phrase such as \txt{Let's think step by step.} before the answer prompt \cite{kojima2022large}. 
This method, commonly called CoT, encourages the model to reveal its reasoning process rather than just providing the final classification. 
We denote this approach as \emph{one-stage zero-shot CoT} QA. 

\subsection{Proposed Framework}
\subsubsection{Two-Stage Reasoning for VQA.}
According to recent theoretical and experimental evidence \cite{abbe2025far}, the Transformer architecture often benefits when a complex task is decomposed into simpler sub-tasks. 
Motivated by this observation, we propose an approach to help the Transformer arrive at a more accurate diagnosis in a VQA setting. The graphical illustration is shown in Figure~\ref{fig:framework}.

\paragraph{Stage 1: Visual Description Extraction.}
In the first stage, we instruct the VLM to generate descriptions on visual features of the input image without directly querying for a diagnosis. 
Concretely, we might prompt the VLM as follows: $v = \mathsf{VLM}\bigl(\bx, q_1 :=$ \txt{Describe visual features detected in the image.}$\bigr).$
By avoiding any diagnosis-related question at this stage, we minimize the risk of bias in identifying and describing relevant image features (See Figure~\ref{fig:reliability} in Section~\ref{sec:experiment} for detailed analysis).

\paragraph{Stage 2: Diagnosis via LLM.}
In the second stage, we provide the visual descriptions $v$ as input to a (potentially different) LLM that produces a final diagnosis. 
For example, we can construct the query: $q_2(v) :=$ \txt{Decide which class best matches the visual features described: 0 (normal) or 1 (pneumonia). **Features:** \{features\}}, where we substitute $\{\texttt{features}\}$ with the previously generated $v$. 
Then, the diagnosis is obtained via $a = \mathsf{LLM}\bigl(q_2(v)\bigr).$

\paragraph{Merits.}
Our framework offers several advantages.
First, it enables more transparent inspection of intermediate outputs, potentially increasing clinicians' trust in automated systems.
Second, it explicitly separates the extraction of raw visual information from the reasoning about clinical decisions, thereby providing flexible combinations of models in each stage.
For example, VLMs are generally specialized in extracting and describing visual features, whereas LLMs excel at understanding instructions and providing reasoned outputs.
Hence, one can choose a VLM best suited for image analysis while selecting an LLM most appropriate for clinical reasoning on the target disease.
Third, because smaller LLMs (e.g., 3B) can still exhibit strong reasoning capabilities, using them for the second stage instead of the VLM used in the first stage (e.g., 11B) can reduce computational overhead (See Figure~\ref{fig:smallmodels} in Section~\ref{sec:experiment} for detailed analysis).

\subsubsection{Enhancing Reliability using Test-time Scaling.}

During inference, conventional approaches generate a sequence of tokens using greedy search---selecting the highest-probability token at each autoregressive step---for GPT-style decoders, which can lead to overly generic or suboptimal responses.
We propose to apply \emph{test-time scaling (TTS)} \cite{yao2023tree, snell2024scaling} to enhance accuracy and reliability.

In the first stage, we apply temperature scaling \cite{guo2017calibration} and sample $N$ outputs from the VLM, rather than generating a single textual output for the visual features: $\{v^{(i)}\}_{i=1}^N = \mathsf{VLM}\bigl(\bx, q_1\bigr)$.
Each $v^{(i)}$ is then passed to the second-stage LLM: $a^{(i)} = \mathsf{LLM}\bigl(q_2(v^{(i)})\bigr)$.
Since we want to parse a clear classification label (e.g., 0 or 1) from the LLM outputs, we follow prior methods \cite{toshniwal2024openmathinstruct,toshniwal2025openmathinstruct} by explicitly instructing the LLM to \txt{Strictly adhere to the format by outputting only the final grade inside \textbackslash boxed\{\} and nothing else.}.

In principle, each generation $v^{(i)}$ can be viewed as a sample drawn from the distribution $p(v \mid \bx)$. The second stage provides $p(\hat{y}\mid v)$, and we approximate
\begin{equation}
p(\hat{y}=1 \mid \bx) = \int p(\hat{y}=1 \mid v)\,p(v \mid \bx)\, dv \approx \frac{1}{N} \sum_{i=1}^{N} \mathbb{I} \Bigl( a^{(i)} = \text{\txt{\textbackslash boxed\{1\}}} \Bigr) .
\end{equation}
where $p(v \mid \bx)$ is the distribution over visual descriptions, $p(\hat{y}\mid v)$ is the likelihood of a specific disease given the descriptions, and $\mathbb{I}$ is an indicator function.
For multi-class classification tasks, this approach can be extended by using one-vs.-all (OvA) or one-vs.-one (OvO) strategies \cite{hastie2009elements}, which we leave for future work.

\section{Results and Discussion} \label{sec:experiment}
\subsubsection{Datasets and Models. }
To evaluate the effectiveness of our framework across modalities and diseases, we conduct experiments using PneumoniaMNIST, PathMNIST, and RetinaMNIST from MedMNIST v2 \cite{yang2023medmnist}.
Pneumonia detection is performed using PneumoniaMNIST, which consists of 390 pneumonia cases and 234 normal cases from frontal X-ray images. 
PathMNIST is utilized for colorectal cancer classification, containing 1,233 cases of colorectal adenocarcinoma epithelium and 741 cases of normal colon mucosa.
Diabetic retinopathy (DR) classification is explored with RetinaMNIST, which includes 226 cases of referable (i.e., non-proliferative or proliferative DR) and 174 normal cases from fundus images.
All images are standardized to a resolution of 224 × 224 pixels.

This study aims to assess the applicability of small general-purpose LLMs that can operate on a single 32GB GPU; we utilize \texttt{Llama-3.2-11B-Vision-Instruct} \cite{touvron2023llama} for the first and second stages.
In addition, we evaluate smaller 1B, 3B, and 8B sizes of \texttt{Llama} text-only models and a medical domain-specific text-only model, \texttt{Med42-v2-8B} \cite{christophe2024med42} for the second stage.
Due to space limits, detailed prompts and codes used in the baseline and proposed frameworks are omitted here but will be shared at publication time.

\input{table/table_comparison}
\subsubsection{Comparison with Baselines.}

For each dataset, we assess the classification performance of standard zero-shot, one-stage zero-shot CoT, and our approach under both single-sample and TTS settings.

Table~\ref{tab:comparison} shows that our method with TTS outperforms or matches the zero-shot baselines.
Furthermore, the proposed TTS strategy consistently enhances classification performance across various methods, including both our approach and the baselines.
Meanwhile, the na\"ive CoT prompting does not significantly improve classification performance and sometimes even degrades it.
In addition, our preliminary experiments indicate that simply combining the prompts used for both stages into a single round (i.e., describe-``and''-diagnose) results in performance analogous to that of one-stage zero-shot CoT.

\input{figure/fig_tts}

We conduct an ablation study to assess the impact of varying the number of TTS samples, $N$, as illustrated in Figure~\ref{fig:scaling}.

Consistent with findings from prior studies, performance follows a power law: as $N$ grows, the predictions become more robust, leading to significant improvements in classification accuracy compared to single-sample approaches.

Notably, even a relatively small $N$ (e.g., 4) yields substantial gains and outperforms the baselines using a single sample. This highlights how diversified visual descriptions enrich the second-stage LLM’s inputs and mitigate reliance on a single, potentially flawed, description.

Our method exhibits low performance when using only a single sample ($N=1$).
This is likely because the first stage does not explicitly dictate the model to focus on visual features related to the specific disease, potentially reducing sample efficiency (i.e., it may attend to non-pneumonia features in the chest X-ray).
However, the subsequent analysis shows that this unconstrained questioning is one of the key factors in enhancing overall reliability.

\input{figure/fig_ablation}

\subsubsection{Generic Prompt Enhances Reliability.}
We examine the impact of unconstrained prompting in Stage 1.
Rather than simply instructing to provide visual descriptions, we \emph{dictate} the model by prompting \txt{Include only features directly associated with identifying pneumonia}, for instance.
We compare the AUC of the unconstrained versus dictated prompting in the first stage and show the result in Figure~\ref{fig:reliability}.

We observe that prompting the VLM directly with diagnosis-related questions that suggest a particular disease, may make the VLM to more likely \emph{hallucinate} the answer and introduce bias (e.g., overdiagnosis).
In contrast, our approach first asks the model to list unbiased visual features by maintaining a neutral, open-ended prompting. This appears to help mitigate such biases and reduce hallucinated findings by separating visual feature extraction from the diagnostic decision.

The observation is particularly compelling as it closely aligns with recent findings from LLM hallucination studies. \cite{johnson2024experts, yadkori2024believe} the theoretical foundation and empirical evidence for \emph{Experts Don't Cheat} concept, suggesting that when a model lacks confidence in a question, it tends to hallucinate answers based on hints in the input, regardless of their factual accuracy.

Similarly, when prompted about a specific disease, VLMs are prone to generate biased responses, especially when they are uncertain about the input image. 
We believe that mitigating this issue and enhancing the robustness of VLMs in safety-critical tasks is a promising direction for future research.

\subsubsection{Ablation on LLM size in the second stage. } 
To explore the method's flexibility and computational efficiency, we replace the original 11B LLM in Stage 2 with smaller ones (1B--8B) for diagnosis.
Figure~\ref{fig:smallmodels} indicates that the performance remains high even for a 3B model, which is comparable to 8B or 11B performance.
For pneumonia diagnosis task, even a 1B model performs reasonably well.

Furthermore, those \texttt{Llama} models achieve results similar to those of a domain-specific fine-tuned model, \texttt{Med42-v2-8B}, suggesting that it encodes an inherent understanding of medical knowledge.

Hence, our method can reduce computational demands without significantly compromising diagnostic performance.

In our preliminary experiments, we tested smaller VLMs, including \texttt{Phi-3.5-vision-instruct}, \texttt{Qwen2.5-VL-3B-Instruct}, and \texttt{DeepSeek-VL2} with 1.0B, 2.8B, and 4.5B, for medical image interpretation in Stage 1. However, these models generated overly generic outputs, failing to capture domain-specific nuances effectively.
Additionally, while the use of huge models (e.g., \texttt{Llama-3.2-90B-Vision-Instruct} could lead to better understanding of the medical image and description, it is constrained by GPU limitations, making deployment impractical in resource-limited settings. 
These challenges highlight the need for future work on exploring smaller yet well-distilled models capable of domain-specific visual understanding.

\section{Conclusion}
We propose a framework that enhances zero-shot medical image diagnosis by integrating multimodal LLM reasoning with test-time scaling. Evaluations across imaging modalities show that our method robustly outperforms baseline approaches, with increased visual description diversity during test-time scaling. Furthermore, the effects of unconstrained prompting and compute savings of our method suggest potential avenues for reducing hallucinations and employing smaller, well-distilled models for advanced visual understanding.

\paragraph{Implications and Future Directions. }
Even with significant gains in AI for medicine, concerns persist due to the inherent ``black-box'' nature of conventional deep learning models. In contrast, LLMs generate text and reasoning reveals the underlying rationale, which helps us understand their logic and mitigate black-box issues. Moreover, integrating LLM-generated reasoning reduces reliance on expensive and scarce human expert annotations, thereby enhancing scalability and cost efficiency in clinical applications and ultimately yielding safer, more effective systems beyond disease classification.

On the other hand, due to the scarcity of high-quality medical annotations required for training verifier or process reward models---often trained upon expensive human annotations or extensive Monte Carlo rollouts---this paper focuses on a simple TTS method.
In the future, more sophisticated reward-model-based approaches could offer promising directions, particularly as medical-domain-specific resources become more readily available.

\bibliographystyle{splncs04}
\bibliography{references}

\end{document}

%% file: notation.tex
\usepackage{amsmath}
\usepackage{amssymb}
\usepackage{bm}

\newcommand*\justify{%
  \fontdimen2\font=0.4em%
  \fontdimen3\font=0.2em%
  \fontdimen4\font=0.1em%
  \fontdimen7\font=0.1em%
  \hyphenchar\font=`\-%
}

\newcommand{\bx}{\bm{x}}

\newcommand{\txt}[1]{``\texttt{\justify #1}''}

%% file: table/table_comparison.tex
\newcommand{\greedy}{Single}
\newcommand{\scaling}{TTS}

\newcommand{\rocauc}{AUC}
\newcommand{\acc}{AP}

\begin{table}[t!]
\centering
{\fontsize{8pt}{10pt}\selectfont
\begin{tabular}{l|cc|cc|cc|cc|cc|cc}
\toprule
 & \multicolumn{4}{c|}{\bf Zero-shot} & \multicolumn{4}{c|}{\bf \makecell{One-stage CoT}} & \multicolumn{4}{c}{\bf \makecell{Describe-then-\\Diagnose (Ours)}} \\
 & \multicolumn{2}{c}{\greedy} & \multicolumn{2}{c|}{\scaling} & \multicolumn{2}{c}{\greedy} & \multicolumn{2}{c|}{\scaling} & \multicolumn{2}{c}{\greedy} & \multicolumn{2}{c}{\scaling} \\
 & \rocauc & \acc & \rocauc & \acc & \rocauc & \acc & \rocauc & \acc & \rocauc & \acc & \rocauc & \acc \\
\midrule
PneumoniaMNIST & 0.495 & 0.621 & 0.737 & 0.790 & 0.530 & 0.643 & \underline{0.779} & \underline{0.831} & 0.517 & 0.634 & \textbf{0.821} & \textbf{0.862} \\
PathMNIST & 0.557 & 0.653 & \underline{0.562} & \underline{0.656} & 0.475 & 0.615 & 0.529 & 0.639 & 0.544 & 0.646 & \textbf{0.653} & \textbf{0.752} \\
RetinaMNIST & 0.607 & 0.625 & \textbf{0.707} & \underline{0.736} & 0.580 & 0.609 & 0.672 & \textbf{0.743} & 0.570 & 0.606 & \underline{0.705} & \underline{0.736} \\
\bottomrule
\end{tabular}
}
\caption{Comparison of our approach against baseline methods---zero-shot and one-stage zero-shot CoT---under a decoding strategy that uses a \textbf{single} sample ($N=1$) and the proposed \textbf{test-time scaling (TTS)} method ($N=16$). TTS provides robust performance gains in zero-shot medical image diagnosis, particularly enhancing our approach with PneumoniaMNIST and PathMNIST. (AUC: area under the receiver operating characteristic curve. AP: area under the precision-recall curve. The \textbf{best} and the \underline{second best} are highlighted.)
}
\label{tab:comparison}
\end{table}

%% file: figure/fig_tts.tex
\begin{figure}[t]
    \centering
    \begin{subfigure}[]{0.32\textwidth}
        \centering
        \includegraphics[width=\linewidth]{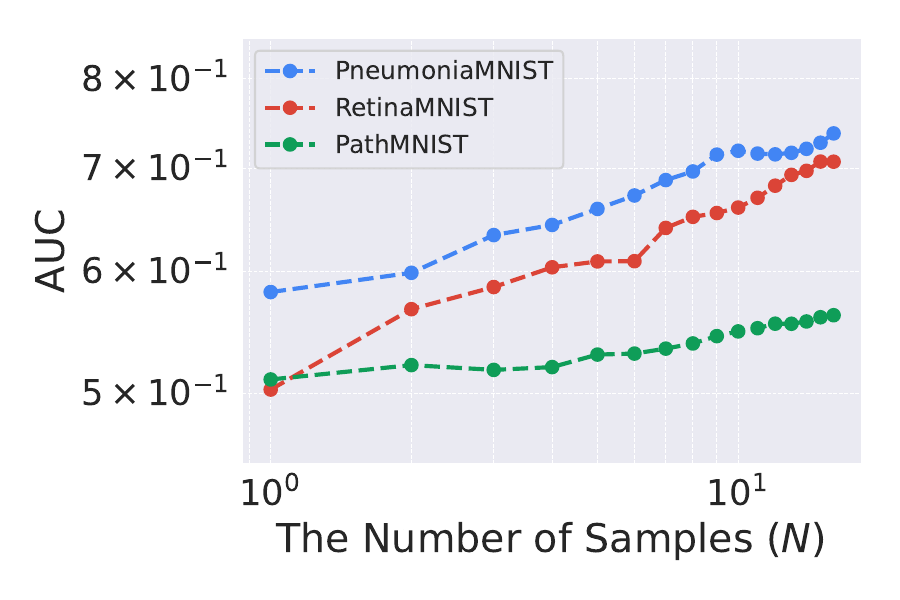}
        \caption{Zero-shot}
    \end{subfigure}
    \begin{subfigure}[]{0.32\textwidth}
        \centering
        \includegraphics[width=\linewidth]{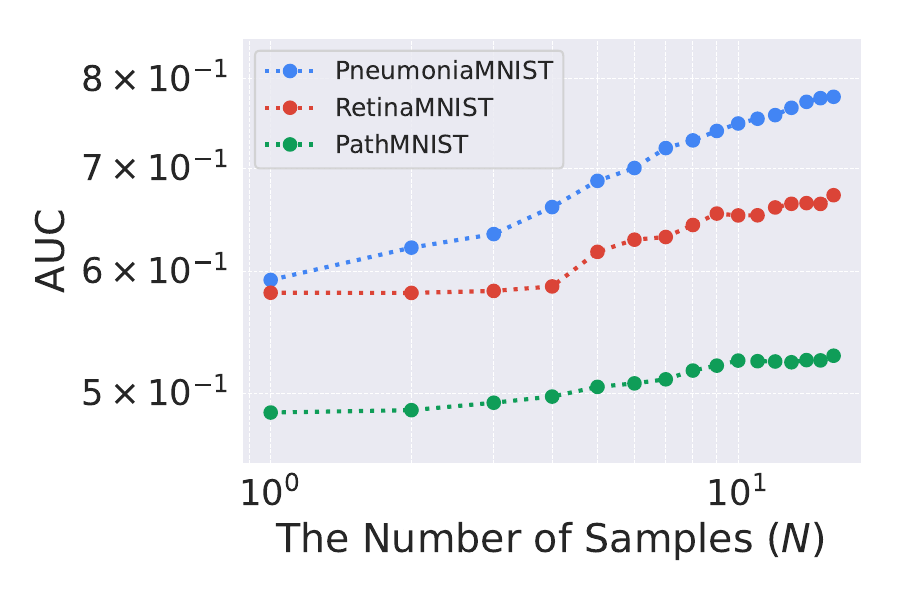}
        \caption{One-stage CoT}
    \end{subfigure}
    \begin{subfigure}[]{0.32\textwidth}
        \centering
        \includegraphics[width=\linewidth]{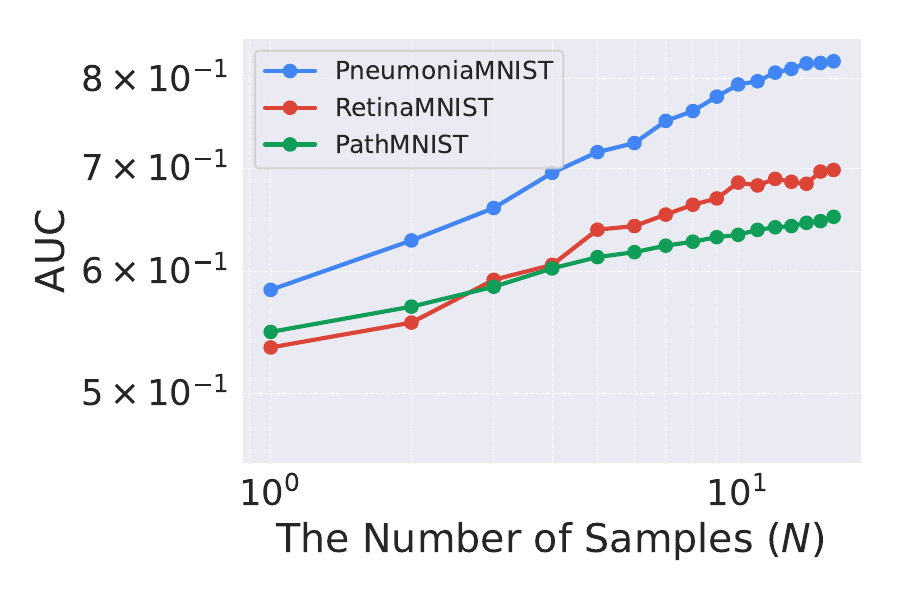}
        \caption{Describe-then-Diagnose}
    \end{subfigure}
    \caption{An ablation study examining the effect of sample size (N) in TTS setting. Increasing the sample size boosts classification performance across different datasets and inference methods---including zero-shot, one-stage CoT, and describe-then-diagnose (ours)---following a power law.}
    \label{fig:scaling}
\end{figure}

%% file: figure/fig_ablation.tex
\begin{figure}[!ht]
    \centering
    \begin{subfigure}[]{0.495\textwidth}
        \centering
        \includegraphics[width=\linewidth]{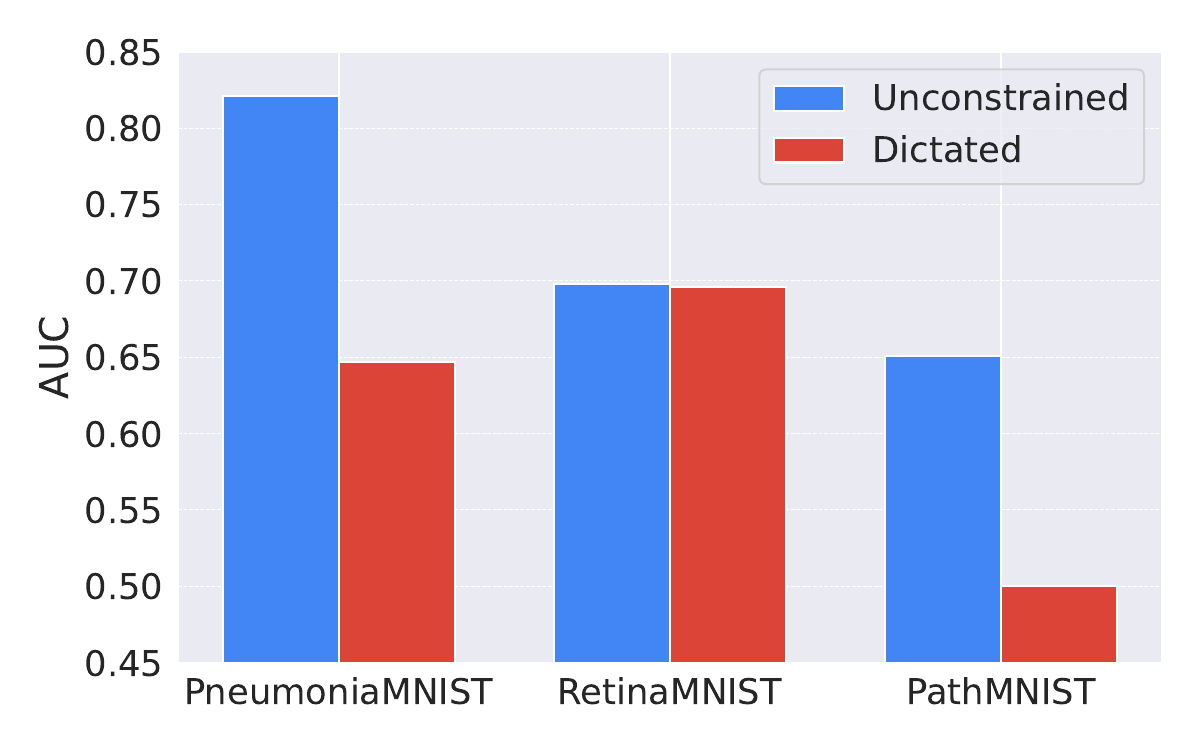} 
        \caption{Bias mitigation in Stage 1}
        \label{fig:reliability}
    \end{subfigure}
    \begin{subfigure}[]{0.495\textwidth}
        \centering
        \includegraphics[width=\linewidth]{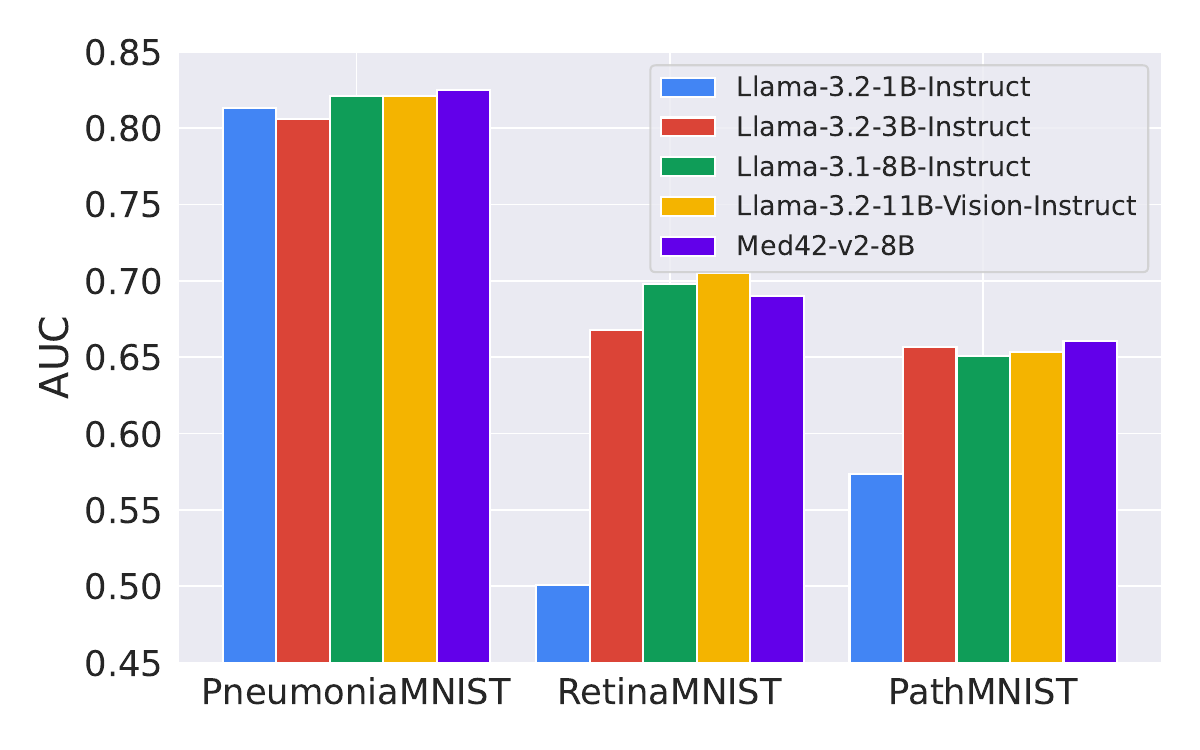} 
        \caption{Compute savings in Stage 2} \label{fig:smallmodels}
    \end{subfigure}
    \caption{Panel (a) illustrates how biased prompting affects diagnostic performance. Direct diagnosis-related, \textbf{dictated} prompts may lead the VLM to hallucinate answers and overdiagnose, whereas neutral, \textbf{unconstrained} prompts that first extract unbiased visual features mitigate bias and improve performance. Panel (b) shows the proposed method's flexibility by replacing the Stage 2 LLM with models of varying sizes. A 3B model performs comparably to an 11B model, demonstrating the benefit of using the two-stage framework. Moreover, 3B achieves performance similar to a domain-specific fine-tuned 8B model.}
\end{figure}

%% file: main.bbl
\begin{thebibliography}{10}
\providecommand{\url}[1]{\texttt{#1}}
\providecommand{\urlprefix}{URL }
\providecommand{\doi}[1]{https://doi.org/#1}

\bibitem{abbe2025far}
Abbe, E., Bengio, S., Lotfi, A., Sandon, C., Saremi, O.: How far can transformers reason? the globality barrier and inductive scratchpad. Advances in Neural Information Processing Systems  \textbf{37},  27850--27895 (2025)

\bibitem{christophe2024med42}
Christophe, C., Kanithi, P.K., Raha, T., Khan, S., Pimentel, M.A.: Med42-v2: A suite of clinical llms. arXiv preprint arXiv:2408.06142  (2024)

\bibitem{cobbe2021training}
Cobbe, K., Kosaraju, V., Bavarian, M., Chen, M., Jun, H., Kaiser, L., Plappert, M., Tworek, J., Hilton, J., Nakano, R., et~al.: Training verifiers to solve math word problems. arXiv preprint arXiv:2110.14168  (2021)

\bibitem{guo2017calibration}
Guo, C., Pleiss, G., Sun, Y., Weinberger, K.Q.: On calibration of modern neural networks. In: International conference on machine learning. pp. 1321--1330. PMLR (2017)

\bibitem{hastie2009elements}
Hastie, T., Tibshirani, R., Friedman, J.H., Friedman, J.H.: The elements of statistical learning: data mining, inference, and prediction, vol.~2. Springer (2009)

\bibitem{jin2021disease}
Jin, D., Pan, E., Oufattole, N., Weng, W.H., Fang, H., Szolovits, P.: What disease does this patient have? a large-scale open domain question answering dataset from medical exams. Applied Sciences  \textbf{11}(14), ~6421 (2021)

\bibitem{jin2019pubmedqa}
Jin, Q., Dhingra, B., Liu, Z., Cohen, W.W., Lu, X.: Pubmedqa: A dataset for biomedical research question answering. arXiv preprint arXiv:1909.06146  (2019)

\bibitem{johnson2024experts}
Johnson, D.D., Tarlow, D., Duvenaud, D., Maddison, C.J.: Experts don't cheat: Learning what you don't know by predicting pairs. arXiv preprint arXiv:2402.08733  (2024)

\bibitem{johnson2023assessing}
Johnson, D., Goodman, R., Patrinely, J., Stone, C., Zimmerman, E., Donald, R., Chang, S., Berkowitz, S., Finn, A., Jahangir, E., et~al.: Assessing the accuracy and reliability of ai-generated medical responses: an evaluation of the chat-gpt model. Research square pp. rs--3 (2023)

\bibitem{kojima2022large}
Kojima, T., Gu, S.S., Reid, M., Matsuo, Y., Iwasawa, Y.: Large language models are zero-shot reasoners. Advances in neural information processing systems  \textbf{35},  22199--22213 (2022)

\bibitem{li2023llava}
Li, C., Wong, C., Zhang, S., Usuyama, N., Liu, H., Yang, J., Naumann, T., Poon, H., Gao, J.: Llava-med: Training a large language-and-vision assistant for biomedicine in one day. Advances in Neural Information Processing Systems  \textbf{36},  28541--28564 (2023)

\bibitem{lightman2024lets}
Lightman, H., Kosaraju, V., Burda, Y., Edwards, H., Baker, B., Lee, T., Leike, J., Schulman, J., Sutskever, I., Cobbe, K.: Let's verify step by step. In: The Twelfth International Conference on Learning Representations (2024), \url{https://openreview.net/forum?id=v8L0pN6EOi}

\bibitem{moor2023medflamingo}
Moor, M., Huang, Q., Wu, S., Yasunaga, M., Dalmia, Y., Leskovec, J., Zakka, C., Reis, E.P., Rajpurkar, P.: Med-flamingo: a multimodal medical few-shot learner. In: Machine Learning for Health (ML4H). pp. 353--367. PMLR (2023)

\bibitem{snell2024scaling}
Snell, C., Lee, J., Xu, K., Kumar, A.: Scaling llm test-time compute optimally can be more effective than scaling model parameters. arXiv preprint arXiv:2408.03314  (2024)

\bibitem{temsah2024openai}
Temsah, M.H., Jamal, A., Alhasan, K., Temsah, A.A., Malki, K.H.: Openai o1-preview vs. chatgpt in healthcare: a new frontier in medical ai reasoning. Cureus  \textbf{16}(10) (2024)

\bibitem{toshniwal2025openmathinstruct}
Toshniwal, S., Du, W., Moshkov, I., Kisacanin, B., Ayrapetyan, A., Gitman, I.: Openmathinstruct-2: Accelerating {AI} for math with massive open-source instruction data. In: The Thirteenth International Conference on Learning Representations (2025), \url{https://openreview.net/forum?id=mTCbq2QssD}

\bibitem{toshniwal2024openmathinstruct}
Toshniwal, S., Moshkov, I., Narenthiran, S., Gitman, D., Jia, F., Gitman, I.: Openmathinstruct-1: A 1.8 million math instruction tuning dataset. Advances in Neural Information Processing Systems  \textbf{37},  34737--34774 (2024)

\bibitem{touvron2023llama}
Touvron, H., Lavril, T., Izacard, G., Martinet, X., Lachaux, M.A., Lacroix, T., Rozi{\`e}re, B., Goyal, N., Hambro, E., Azhar, F., et~al.: Llama: Open and efficient foundation language models. arXiv preprint arXiv:2302.13971  (2023)

\bibitem{tu2024medpalm}
Tu, T., Azizi, S., Driess, D., Schaekermann, M., Amin, M., Chang, P.C., Carroll, A., Lau, C., Tanno, R., Ktena, I., et~al.: Towards generalist biomedical ai. Nejm Ai  \textbf{1}(3),  AIoa2300138 (2024)

\bibitem{uesato2022solving}
Uesato, J., Kushman, N., Kumar, R., Song, F., Siegel, N., Wang, L., Creswell, A., Irving, G., Higgins, I.: Solving math word problems with process-and outcome-based feedback. arXiv preprint arXiv:2211.14275  (2022)

\bibitem{want2024shepherd}
Wang, P., Li, L., Shao, Z., Xu, R.X., Dai, D., Li, Y., Chen, D., Wu, Y., Sui, Z.: Math-shepherd: Verify and reinforce llms step-by-step without human annotations. CoRR  \textbf{abs/2312.08935} (2023), \url{https://doi.org/10.48550/arXiv.2312.08935}

\bibitem{wei2022chain}
Wei, J., Wang, X., Schuurmans, D., Bosma, M., Xia, F., Chi, E., Le, Q.V., Zhou, D., et~al.: Chain-of-thought prompting elicits reasoning in large language models. Advances in neural information processing systems  \textbf{35},  24824--24837 (2022)

\bibitem{yadkori2024believe}
Yadkori, Y.A., Kuzborskij, I., Gy{\"o}rgy, A., Szepesv{\'a}ri, C.: To believe or not to believe your llm. arXiv preprint arXiv:2406.02543  (2024)

\bibitem{yang2023medmnist}
Yang, J., Shi, R., Wei, D., Liu, Z., Zhao, L., Ke, B., Pfister, H., Ni, B.: Medmnist v2-a large-scale lightweight benchmark for 2d and 3d biomedical image classification. Scientific Data  \textbf{10}(1), ~41 (2023)

\bibitem{yao2023tree}
Yao, S., Yu, D., Zhao, J., Shafran, I., Griffiths, T., Cao, Y., Narasimhan, K.: Tree of thoughts: Deliberate problem solving with large language models. Advances in neural information processing systems  \textbf{36},  11809--11822 (2023)

\end{thebibliography}
